\documentclass[a4paper]{siamonline1116}

\usepackage{graphicx} \usepackage{amsfonts} 
\usepackage{algorithm} 
\usepackage[noend]{algpseudocode} 

\newcommand{\tld}[1]{\widetilde{#1}}

\DeclareMathOperator{\card}{card}

\newcommand{\naturals}{\ensuremath{\mathbb{N}}}
\newcommand{\reals}{\ensuremath{\mathbb{R}}}
\newcommand{\stoch}{\mathcal{M}} 
\newcommand{\Ell}{\mathcal{L}} 

\renewcommand{\epsilon}{\varepsilon}

\newcommand{\coloneq}{\mathrel{\mathop:}=}

\newcommand{\given}{\mid}

\newcommand\blfootnote[1]{%
  \begingroup
  \renewcommand\thefootnote{}\footnote{#1}%
  \addtocounter{footnote}{-1}%
  \endgroup
}

\title{\LARGE \bf
Bayesian Classifier for Route Prediction with Markov Chains\thanks{This work has been conducted within the ENABLE-S3 project that has received funding from the ECSEL Joint Undertaking under grant agreement no 692455. This joint undertaking receives support from the European Union Horizon 2020 research and innovation programme and Austria, Denmark, Germany, Finland, Czech Republic, Italy, Spain, Portugal, Poland, Ireland, Belgium, France, Netherlands, United Kingdom, Slovakia, Norway.}
}

\author{Jonathan P.\ Epperlein\thanks{IBM Research -- Ireland 
		\texttt{\{jpepperlein,julien.monteil,sergiy.zhuk\}@ie.ibm.com}} \and
		Julien Monteil\footnotemark[2] \and Mingming Liu\thanks{University College Dublin, Ireland
		\texttt{yingqi.gu@ucdconnect.ie}, \texttt{robert.shorten@ucd.ie}, \texttt{mingming.liu@ibm.com}} \and 
		Yingqi Gu\footnotemark[3] \and Sergiy Zhuk\footnotemark[2] \and Robert Shorten\footnotemark[2]~\footnotemark[3]
}

\begin{document}
\maketitle
\thispagestyle{empty}
\begin{abstract}
We present here a general framework and a specific algorithm for predicting the destination, route, or more generally a pattern, of an ongoing journey, building on the recent work of~\cite{JulienRoutePred}. In the presented framework, known journey patterns are modelled as stochastic processes, emitting the road segments visited during the journey, and the ongoing journey is predicted by updating the posterior probability of each journey pattern given the road segments visited so far.
In this contribution, we use Markov chains as models for the journey patterns, and consider the prediction as final, once one of the posterior probabilities crosses a predefined threshold. Despite the simplicity of both, examples run on a synthetic dataset demonstrate high accuracy of the made predictions.
\end{abstract}
\section{INTRODUCTION}\label{sec:intro}
\blfootnote{
\rule{0pt}{3ex}
\copyright 2018 IEEE.
Personal use of this material is permitted. Permission from IEEE must be 
obtained for all other uses, in any current or future media, including reprinting/republishing this material for advertising or promotional purposes,
creating new collective works, for resale or redistribution to servers or lists, or reuse of any 
copyrighted component of this work in other works.}
Understanding driver intent is a prerequisite for a personalised driving experience with features such as personalised risk assessment and mitigation, alerts, advice, automated rerouting, etc. One of the most important driver intentions is the destination of the journey and the route to get there. Furthermore, in particular for hybrid vehicles, knowledge of the route ahead of time can be used to optimise the charge/discharge schedule, e.g.~\cite{Deguchi2004} find improvements in fuel economy of up to 7.8\%.

In the data used in~\cite{FroehlKrumm}, 60\% of trips are repeated and hence predictable from the driving history; \cite{Song1018} suggests that in the more general setting of user mobility, more than 90\% of a user's trajectories are ``potentially predictable.'' The combination of value and feasibility has sparked much research in this direction.
While some algorithms rely on GPS trajectories only -- e.g.\ \cite{FroehlKrumm} computes geometric similarity between curves obtained from cleaned and filtered raw GPS data -- other approaches, such as~\cite{JulienRoutePred,SimmonsBrowningZhangEtAl2006,Krumm2008}, where Markov chains and hidden Markov models are built  on the road network to estimate the most likely routes and destinations, next turns, or trip clusters in a broader sense, require map-matching to first map GPS traces to links in the road network.

{The present paper falls into that latter category, as} it builds on and extends the recent work of~\cite{JulienRoutePred} and contributes to a novel and flexible approach to the important problem of driver intent prediction. It is structured as follows: after introducing some notation relating to probability, stochastic processes, and Markov chains in the next section, we then show in Section~\ref{sec:approach} how trips can be modelled as outputs of stochastic processes to obtain an estimate of the posterior probabilities of each known journey pattern. The algorithm resulting if Markov chains are used as the stochastic process models is described in detail in Section~\ref{sec:pred}, and Section~\ref{sec:exm} provides some experimental validation. We close with some possible extensions and improvements, and the observation that~\cite{JulienRoutePred} also fits into the presented framework, if the stochastic process model is chosen to be a naive Bayes model instead of Markov chains.
%
\section{NOTATION}\label{sec:prelim}
We \emph{should} write $P(W=w)$ for the probability of the event that a realisation of the discrete random variable $W$ equals $w$, and $P(W=w
\given V=v)$ the probability of that same event $W=w$ given that the event $V=v$ occurred. However, for convenience we will most of the time write
$P(w \given v)$ instead of $P(W=w \given V=v)$, when it is clear from the context, what is meant.
For a set of parameters $\mu$ parametrising a probability distribution, the notation $P(W=w \given \mu)$ is taken to denote the probability of the event $W=w$ if the parameters are set to $\mu$. 

We let $\naturals=\{0,1,2,\dotsc\}$ denote the natural numbers, and for $N\in\naturals\setminus\{0\}$, $[N]\coloneq\{1,\dotsc,N\}$. Matrices will be denoted by capital letters, their elements by the same letter in lower case, and the set of $n\times n$ row-stochastic matrices, i.e.\ matrices with non-negative entries such that every row sums up to 1, by $\stoch^n$.
We denote the cardinality of a set $S$, i.e.\ the number of its elements, by $\card S$. All cardinalities here will be finite.

A \emph{stochastic process} is a sequence of random variables $\{X_t\}$ indexed by $t$, which often denotes time. For $t\in\naturals$, we call a sequence $(x_0,x_1,\dotsc,x_T)$ of realisations of the random variables $X_t$ a \emph{trajectory} of the process; $P(x_0,x_1,\dotsc,x_T \given \mu)$ is the \emph{probability} of the trajectory given that the stochastic process is parametrised by $\mu$,\footnote{e.g.\ if $X_t$ corresponds to an unfair coin flip, $\mu$ would correspond to the probability $p$ of heads} whereas interpreted as a function of $\mu$, it is the \emph{likelihood} of the parameters being equal to $\mu$.

Specifically, we will use \emph{Markov chains}, which are stochastic processes with $t\in\naturals$ and $X_t\in[N]$ that are completely defined by a \emph{transition (probability) matrix} $A\in\stoch^N$ and a vector $\pi\in\reals^N$ of initial probabilities. Then, $P(X_0=i)=\pi_i$ and $P(X_t=j \given X_{t-1}=i)=a_{ij}$ characterises the process.
Note that the ``next state'' $X_{t}$ depends \emph{only} on the current realisation $x_{t-1}$ and not on the past; this is also know as the Markov property. This corresponds to a directed graph with $N$ nodes and weights $a_{ij}$ on the edge from node $i$ to $j$. The stochastic process defined by the Markov chain then corresponds to an agent being initialised on some node $i$ according to $\pi$ and  making every decision where to go next according to the weights on the edges leading away from it.

%
\section{PROBLEM SETTING AND FRAMEWORK}\label{sec:approach}
From a driver's history $H$ of trips taken in the past, we want to learn a predictive model, allowing us to identify properties (such as destination or specific route) of a currently ongoing trip as soon into the trip as possible.
The history is 
\begin{multline*}
	H = \Bigl\{  T_1: \{  (t^1_1,t^1_2,\dotsc,t^1_{L_1}   ),(r^1_1,r^1_2,\dotsc,r^1_{L_1}) \}, \dotsc,\\ 
			  T_{N_H} : \{  (t^{N_H}_1,t^{N_H}_2,\dotsc,t^{N_H}_{L_{N_H}}   ),( r^{N_H}_1,r^{N_H}_2,\dotsc,r^{N_H}_{L_{N_H}}) \}    \Bigr\},
\end{multline*}
where each \emph{trip} $T_i$ has length $L_i$ and consists of sequences $(t^i_1,\dotsc )$ and $(r^i_1,\dotsc )$ of \emph{time stamps} and \emph{road segments}. The road segments are identified by their OpenStreetMap (OSM) way IDs~\cite{wiki:way}, which implies that map matching has been performed on the raw GPS trajectories. We'll return to that point in Sec.~\ref{ssec:data}. Let 
\[
	\Ell=\bigcup\nolimits_{i=1}^{N_H} \bigcup\nolimits_{t=1}^{L_i} \{r^i_t\} \qquad \text{ and } \qquad N=\card\Ell
\]
be the set and number of all road segments ever visited.

Each trip in $H$ belongs to a \emph{cluster} $C_k$, where the cluster encodes the journey pattern, destination, or more generally ``properties'' of the trip. From now on, we shall use the more generic term ``cluster,'' as defined in~\cite{JulienRoutePred}; see also there for further explanations. A cluster could be as coarse as a collection of all trips $T_i$ with the same destination (defined as e.g.\ the last road segment of a trip), hence encoding only the property of destination, or more fine-grained, by defining a measure of similarity between trips and clustering according to those similarities; for instance trips along the ``scenic route to work'' and using the ``fastest route to work'' would then belong to different clusters despite sharing the same destination.
More details are again postponed until the computational examples are described in Sec.~\ref{sec:exm}. Whichever way it is obtained, let us define this set of clusters by
\begin{equation}
	C = \{C_1,\dotsc, C_{N_C} \},
\end{equation}
and assume that $C_k$ is a partition of $H$, i.e.\ every trip belongs to \emph{exactly one} cluster. 
We can then state the problem more precisely as:\\[.5ex]
\centerline{\begin{fbox}{\parbox{0.8\columnwidth}{Given an ongoing trip $T:\{ (t_1,\dotsc,t_L),(r_1,\dotsc,r_L) \}$, decide what cluster~$C_k$  trip $T$ belongs to.}}
\end{fbox}}\\[.5ex]
The proposed framework consists of a model for each cluster, providing a likelihood function $P( T \given C_k)$, a prior probability $P(C_k)$, the Bayesian update to the posterior probabilities $P(C_k \given T)$, and the criterion by which the prediction is made. The next sections elaborate on these parts.
%
\subsection{Journeys as Stochastic Processes}\label{ssec:stochproc}
Inspired by classical single-word speech recognition algorithms -- see e.g.\ the classic survey~\cite{Rabiner1989} -- where words are modelled as stochastic processes (for speech recognition frequently hidden Markov models), emitting sequences of vectors of spectral (and/or temporal) features derived from the acoustic speech signal, and the word corresponding to the stochastic process with the highest likelihood of having generated the present sequence is returned as the result, we propose to model journey patterns as stochastic processes emitting road links.

The choice of the type of stochastic process is free, there can even be different kinds for different clusters; once a type is chosen, the parameters for each cluster $C_k$ have to be estimated from the trips in $H$ belonging to $C_k$. In other words, a model for each cluster is ``trained'' on the available history. The outcome of this training process is a mapping $(T,C_k)\mapsto P(T\given C_k)$, i.e.\ a function that allows us to evaluate the likelihood for each cluster of it having produced the currently available sequence of road segments (and time stamps). Our choice of stochastic process here will be Markov chains, which are particularly easy to train and evaluate; the details are given in Sec.~\ref{ssec:MC}.
%
\subsection{Prior Probabilities}\label{ssec:prior}
Before a trip even started, there might already be a high probability of it belonging to a certain cluster: if you have Aikido class on Wednesdays at 19:00 o'clock, and the current trip started on a Wednesday at 18:30 o'clock, there's a very high probability that the trip's destination will be your dojo. More generally, all additional information available about the trip, such as the current day of the week, the weather, current public events etc.\ form the \emph{context} of the trip, and from the context, we can estimate the \emph{prior probabilities}  $P(C_k)$ of each $C_k$ without any trip information yet available. In the absence of context, we can set $P(C_k)=\card\{i\given T_i\in C_k\}/N_C$, i.e.\ make the prior proportional to how many trips in $H$ belonged to $C_k$, or even simply set $P(C_k)=1/N_C$ for all $j\in[N_C]$. 
%
\subsection{Bayesian Updates}\label{ssec:bayes}
Bayes' law relates the quantity we are ultimately interested in -- the probability of the current trip belonging to $C_k$ given what we already know about the trip, i.e.\ $P(C_k \given T)$ -- to the quantities estimated from $H$ -- the likelihood $P(T\given C_k)$ and the prior $P(C_k)$ -- by
\begin{equation} \label{eq:bayes}
	P(C_k\given T) = \frac{P(T\given C_k) P(C_k)}{P(T)}.
\end{equation}
A subtle point concerns the normalisation $P(T)$, the probability of the trip observed so far with no further assumptions on its nature: by computing the numerators $P(T\given C_k) P(C_k)$ and then normalising them to sum up to one, i.e.\ by imposing
\[
	\sum\nolimits_{C_k \in C} P(C_k \given T) = 1,
\]
we implicitly assume that every trip is indeed in one of the already known clusters. As a simple fix, we could introduce the probability of any trip not belonging to any known cluster as a constant $P(\neg {C})$, in which case we'd have $\sum_{C_k \in C} P(C_k \given T) = 1-P(\neg C)$, simply a normalisation to a different constant. We proceed without accounting for unknown clusters, but keep this implicit assumption in mind.
%
\subsection{Stopping Criterion}\label{ssec:stop}
If, when, and how the updates should be stopped and the final prediction announced depends on the application. If the goal is route planning, we need to be reasonably certain of the unique destination before planning a route. On the other hand, if the goal is identifying risks along the remaining journey, it is sufficient to narrow down the set of possible routes and check for risks along all of them; predictions can be made continuously in this scenario. Other applications might call for other measures.

Here, we apply the simple criterion that as soon as one of the clusters' posterior probabilities exceeds a threshold $P(C\given T)>1-\alpha$, prediction stops and returns cluster $C$ as the result, which should work well for the first application.
%
%
\section{CLUSTER PREDICTION ALGORITHM}\label{sec:pred}
In order to derive a concrete algorithm for cluster prediction, a choice of statistical process model has to be made, and the models have to be trained. In this section, we describe this for the case of Markov chains.
%
\subsection{Modelling Clusters as Markov Chains}\label{ssec:MC}
Modelling clusters by Markov chains is, of course, a simplification of reality, but as we shall see it leads to a computationally very tractable algorithm and performs very well in our computational experiments in the next section. The simplifying assumption is the ``Markov assumption'': if the current trip belongs to a cluster $C_k$, the probability distribution of the next road segment\footnote{The capital ``$R$'' is used because it denotes a random variable.}$R_{t+1}$ depends \emph{only} on the current road segment $r_t$. 

More formally, the state space of the Markov chain is $[N]$, where $N$ is the total number of road segments in the road network under consideration. Every trip $T$, or rather its sequence $(r_1,\dotsc,r_L)$ of road segments, then corresponds to a trajectory of the Markov chain.
If a trip belongs to a certain cluster $C_k$, e.g.\ ``scenic route from home to work,'' and if the trip so far has been $r_0,\dotsc,r_t$, then, in full generality, the probability distribution of the next road segment, given all that is known at $t$, is $P(R_{t+1}=r_{t+1} \given r_0,\dotsc,r_t,C_k)$. In modelling this as a Markov chain, we are imposing that 
\begin{equation}
	P(R_{t+1}=j \given r_0,\dotsc,r_{t-1},R_t=i,C_k) =
					 P(R_{t+1}=j \given R_t=i,C_k) = a^k_{ij}.
\end{equation}
Then, $A^k\in\stoch^N$ is the transition probability matrix of cluster $C_k$, and $a^k_{ij}$ is the probability to turn into road $j$ from road $i$, if the current trip is in cluster $C_k$.

The transition probabilities are estimated by
\begin{multline} \label{eq:MLA}
	a^k_{ij} = \frac{ \sum_{T_m \in C_k} \card\{ t \given r^m_t=i, r^m_{t+1}=j  \}      }{ \sum_{T_m \in C_k} \card\{ t \given r^m_t=i, t=1,\dotsc,L_m-1 \}       }     \\[1ex]
	=\frac{\text{number of transitions from $i$ to $j$ in  $C_k$}}{\text{number of times $i$ is transitioned from in $C_k$}}.
\end{multline}
This very intuitive estimate is in fact the maximum-likelihood (ML) estimate, see e.g.~\cite{AndersonGoodman1957}. If a road segment $i$ never appears (or more precisely, is never transitioned from) in any trip in $C_k$, then~\eqref{eq:MLA} cannot be applied; for now, we can just set these $a^k_{ij}$ to 0, and return to this issue in Sec.~\ref{ssec:prank}.

For the initial probabilities $\pi^k_i = P(R_0=i\given C_k)$, the (ML) estimate is 
\begin{equation}\label{eq:MLpi}
	\pi^k_i = \frac{  \card\{ m \given r^m_0=i \text{ and } T_m\in C_k  \}      }{  \card\{ m \given T_m\in C_k  \}      },
\end{equation}
but this assumes that the prediction task always starts at the beginning of a trip; if for whatever reason the first few links of a trip are missing in the data, this might lead to a failure of the prediction. E.g.\ by choosing 
\[
	\pi^k_i = \begin{cases} \frac{1}{\card\{ j \given r^m_t=j \text{ for any } t \text{ and any } T_m\in C_k\} }& \text{if $r^m_t=i$ for any $t$ and any $T_m\in C_k$} \\
					    0 & \text{else, }
			\end{cases}
\]
i.e.\ making the initial probability uniform over all road segments that appear in the cluster, or even setting $\pi^k_i=1/N$ for all $i\in\Ell$, we avoid this problem and allow for prediction to start during a trip; we thus treat $\pi^k$ somewhat heuristically as a design parameter. 
The likelihood function is then given by
\begin{equation}\label{eq:MClik}
	P( r_1, r_2, \dotsc, r_L \given C_k) = \pi^k_{r_1} \prod_{t=2}^{L} a^k_{r_{t-1},r_{t}}, 
\end{equation}
or recursively by
\begin{equation}\label{eq:MClikrec}
	P( r_1, r_2, \dotsc, r_L \given C_k) = P( r_1, r_2, \dotsc, r_{L-1} \given C_k)  a^k_{r_{L-1},r_{L}}.
\end{equation}

\begin{algorithm}[tbh]
\begin{algorithmic}[1]
	\Procedure{Training}{$H$, $C$, $\epsilon>0$}
		\ForAll{$C_k \in C$}
			\State $(A^k,\pi^k) \gets$ by~\eqref{eq:MLA}, \eqref{eq:MLpi}, \eqref{eq:MLAprank}
		\EndFor
	\EndProcedure
	\Procedure{Prediction}{$A^k, \pi^k, P(C_k) \,\forall C_k\in C, \alpha, \bar{\epsilon}$}
		\Statex \emph{Initialization}
		\State $r_\text{to} \gets r_0$, $r_\text{from} \gets $ None
		 \ForAll{$k=1,\dotsc,N_C$}
		 	\State $\ell_k \gets \pi^k_{r_0}$, $P_k \gets \ell_k P(C_k) /\sum_{j=1}^{N_C} P_j P(C_j)$
		\EndFor
		\Statex \emph{Prediction loop}
		\While{$P_k\leq 1-\alpha \;\forall k$}
			\If{trip is finished}
				\State \textbf{return} None
			\Else
            			\State wait until new segment $r_\text{new}$ is received 
            			\State $r_\text{from}\gets r_\text{to}$, $r_\text{to} \gets r_\text{new}$
            			\ForAll{$k=1,\dotsc, N_C$}
            				\If{$r_\text{from} \notin\Ell \vee r_\text{to}\notin \Ell$}
            					\State $\ell_k \gets \ell_k \cdot \bar{\epsilon}$				
            				\Else
            					\State $\ell_k \gets \ell_k \cdot a^k_{r_\text{from}\, r_\text{to}}$\label{algline:prank_here}
            					\Comment{see Eq.~\eqref{eq:MClikrec}}
            				\EndIf				
            				\State $\bar{P}_k \gets \ell_k \cdot P(C_k)$ \Comment{Bayesian update}
            			\EndFor		
            			\ForAll{$k=1,\dotsc, N_C$}
            				\State ${P}_k \gets \bar{P}_k/\sum_{j=1}^{N_C} \bar{P}_j$  \Comment{Normalization}
            			\EndFor		
			\EndIf
		\EndWhile
		\State \textbf{return} $\arg\max_k P_k$
	\EndProcedure
\end{algorithmic}
\caption{Cluster prediction using Markov chains}
\label{alg:clustpred}
\end{algorithm}
%
\subsection{Unseen Transitions and Unseen Road Segments}\label{ssec:prank}
When a transition that has never occurred in the training data occurs in the current trip to be predicted, the algorithm described so far will break down, because the likelihood $P( r_1, r_2, \dotsc, r_L \given C_k)$ will drop to $0$ for all clusters $C_k$, and hence the posteriors $P( C_k \given T)$ will be undefined as $0/0$. This will be a rather common situation: the data is not perfect, hence segments visited in reality could be missing in the data, GPS data could be mapped to the wrong way ID, small detours could take the driver along never before visited roads, etc. Similar to the PageRank algorithm~\cite{Page1999} and as in~\cite{JulienRoutePred}, we address this problem by introducing a small $\epsilon>0$ and adding it to \emph{each} transition probability (except self-loops, i.e.\ transitions $i\rightarrow i$), even the ones never observed. The matrices $A^k$ then have to be re-normalized to obtain stochastic matrices again, however now every probability $a^k_{ij}\geq \epsilon/(1+(N-1)\epsilon)>0$. Formally, for all $i\neq j$:
\begin{align} \label{eq:MLAprank}
	\tld{a}^k_{ij} &= a^k_{ij}+\epsilon & a^k_{ij} &= \frac{\tld{a}^k_{ij}}{\sum_m \tld{a}^k_{im}}.
\end{align}

If the prediction algorithm receives a road segment that has never been seen, i.e.\ if $r_L\notin\Ell$, this is treated in much the same way by extending the likelihood function, or rather the transition probability matrices $A^k$, to assign a minimum probability $\epsilon/(1+(N+1)\epsilon)$ to transitions to or from unseen links. This can be addressed very easily in the code by inserting an \textbf{if} statement before updating the likelihood, but also formally: we add a state $u$ for ``unseen segment'' to $\Ell$ to obtain $\Ell^u=\Ell\cup\{u\}$, and add a column and row equal to $\bar{\epsilon}\coloneq\epsilon/(1+(N+1)\epsilon)$ to every $A^k$ (note that a transition from $u$ to $u$ is now allowed, as it corresponds to two previously unseen segments in sequence, and not necessarily a repetition of the current segment). Every link with an ID not in $\Ell$ is then mapped to $u$ before the likelihood is computed.

Pseudocode of the full algorithm, including the modifications described here, is shown as Algorithm~\ref{alg:clustpred}.
%
%
\section{COMPUTATIONAL EXPERIMENTS}\label{sec:exm}
We now use the dataset in~\cite{JulienRoutePred} to test the proposed algorithm; a short description is given below, for more details, see~\cite{JulienRoutePred}.
\begin{figure}
	\centering\includegraphics[width=.9\textwidth]{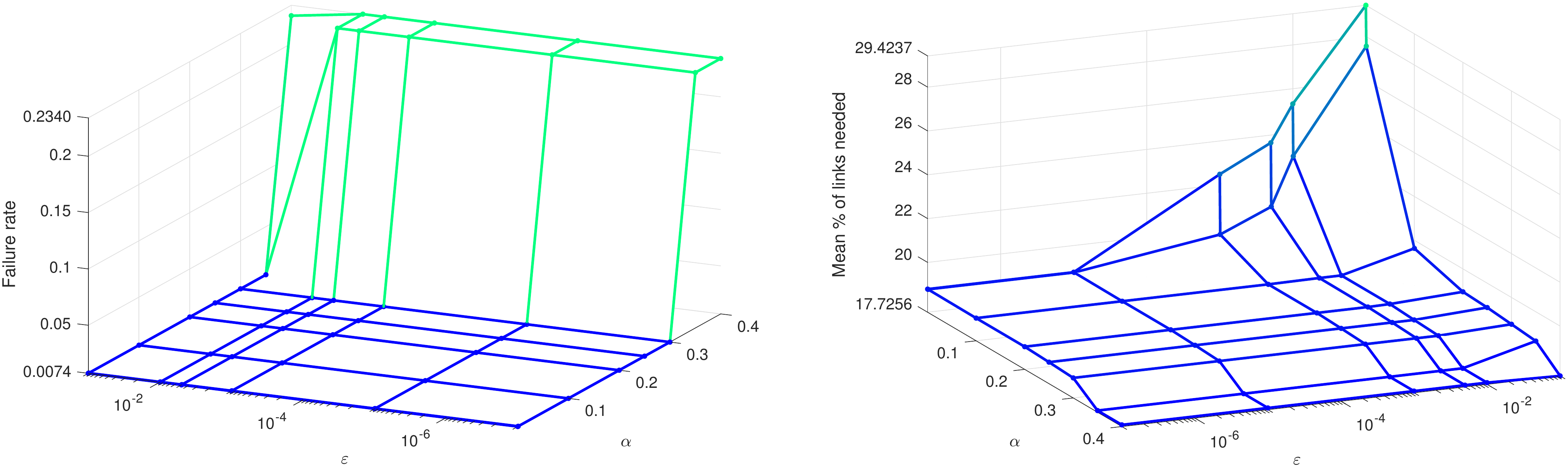}
	\caption{Failure rates and mean percentage of links needed to make a prediction after 8 rounds of 50-50 cross validation. Note the different view angles of the two axes.}
	\label{fig:results}
\end{figure}
%
\subsection{Data}\label{ssec:data}
Seven origins and destinations across Dublin were selected, representing typical points of interest such as ``home,'' ``work,'' ``childcare,'' etc. This yields a total of 21 possible origin/destination pairs, for 17 of which up to 3 distinct routes were generated. These routes were then fed into the microscopic traffic simulator SUMO~\cite{SUMO2012} to generate a total of $N_H=781$ trips in the form of timestamps and longitude/latitude coordinates. To simulate real GPS data, uniformly distributed noise (on a disk of radius 10m) was added to each point. 

To prepare the data in the form of $H$ for our algorithm, the sequence of GPS points needs to be converted into a sequence of way IDs, which was done using the Map Matching operator of IBM Streams Geospatial Toolkit. Subsequently, duplicates were removed (i.e.\ if more than one consecutive GPS point was mapped to the same road segment, only the first instance was kept).
%
\subsection{Clustering}\label{ssec:cluster}
As in~\cite{JulienRoutePred}, we consider two types of clusters:
\begin{itemize}
	\item \emph{by Origin/Destination:} Two trips belong to the same cluster, iff they have the same origin and destination (as defined by proximity on their final GPS coordinates). This results in $N_C=17$ clusters.
	\item \emph{by Route:} Similarity between two trips is measured by the ratio of shared road segments between the two trips and the total number of road segments in both trips. Hierarchical clustering is then performed, and the dissimilarity threshold is chosen to be 0.3, yielding $N_C=30$ clusters.
\end{itemize}
%
\subsection{Prediction Results}\label{ssec:res}
\begin{figure}[tbh]
	\centering
	\parbox[b]{.48\textwidth}{
		\centering
		\includegraphics[width=.48\textwidth]{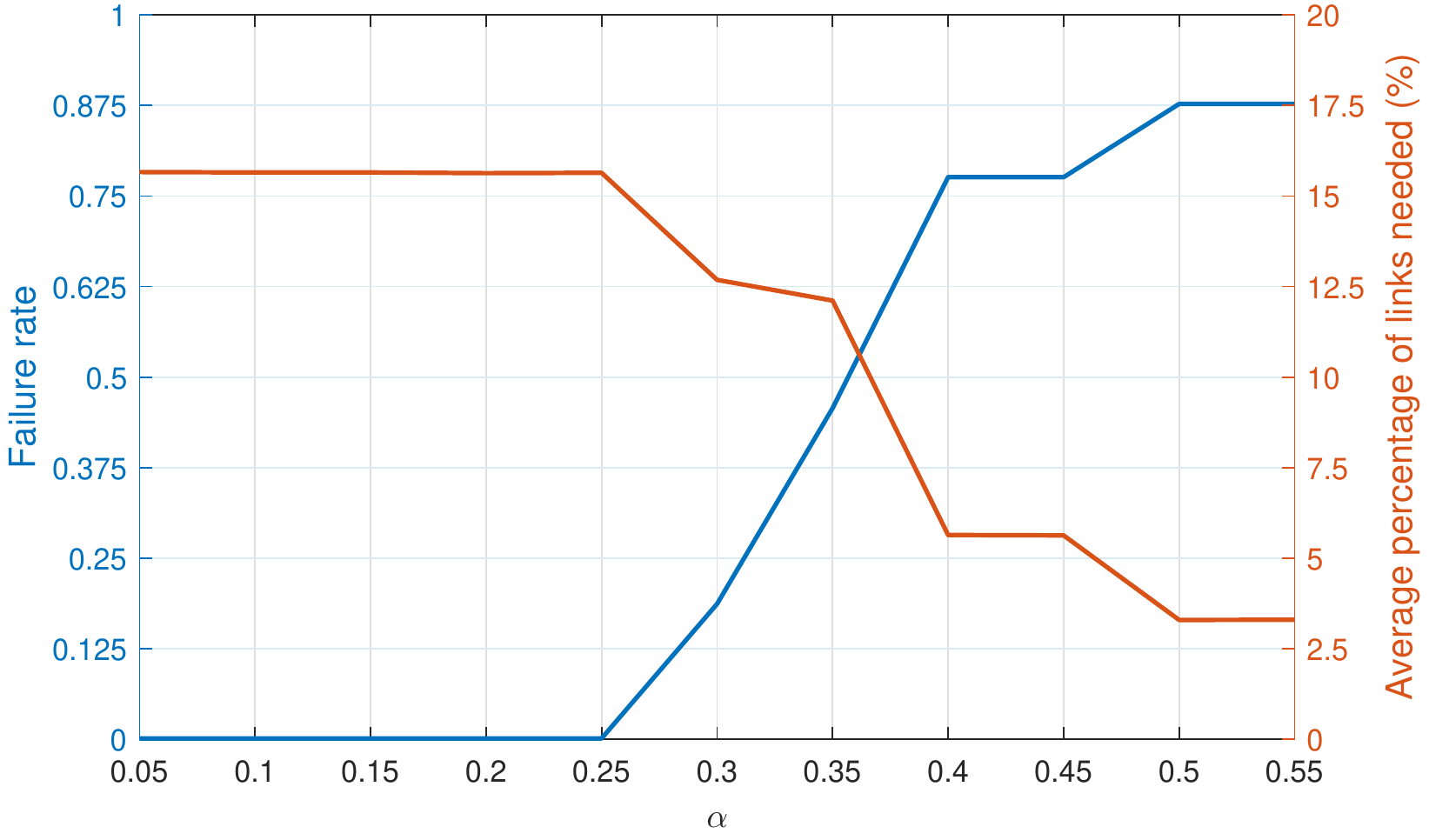}
		\caption{Failure rate and fraction of links needed for prediction vs confidence level $\alpha$ for fixed $\epsilon = 10^{-6}$ if clustering by Origin/Destination.}
		\label{fig:leave1:alpha:OD}
		}
		\hfill{}
	\parbox[b]{.48\textwidth}{
		\centering
  		\includegraphics[width=.48\textwidth]{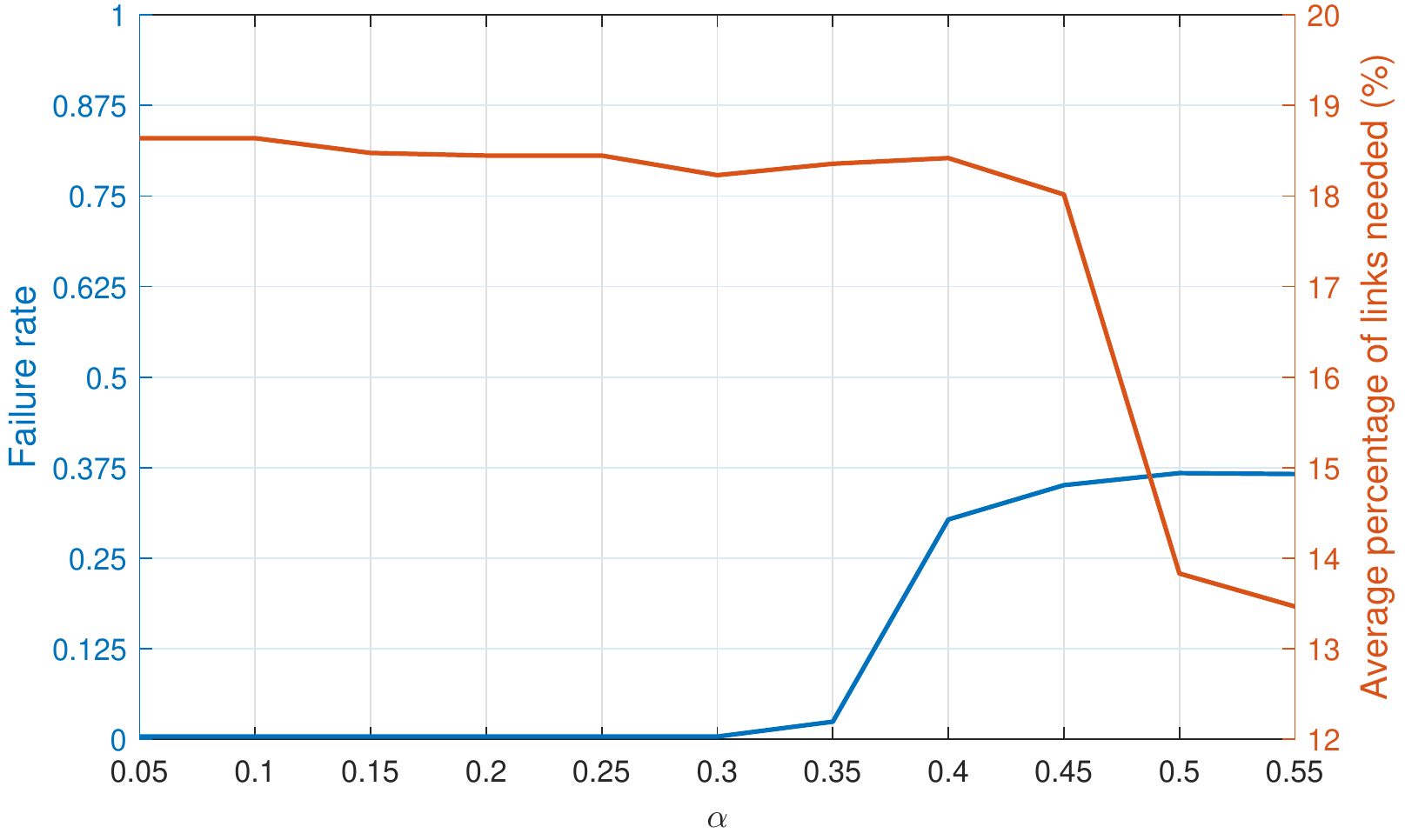}
		\caption{Failure rate and fraction of links needed for prediction vs confidence level $\alpha$ for fixed $\epsilon = 10^{-6}$ if clustering by Route.}
   		\label{fig:leave1:alpha:Route}
		}
\end{figure}
\begin{figure}[tbh]
\centering
		\includegraphics[width=.8\columnwidth]{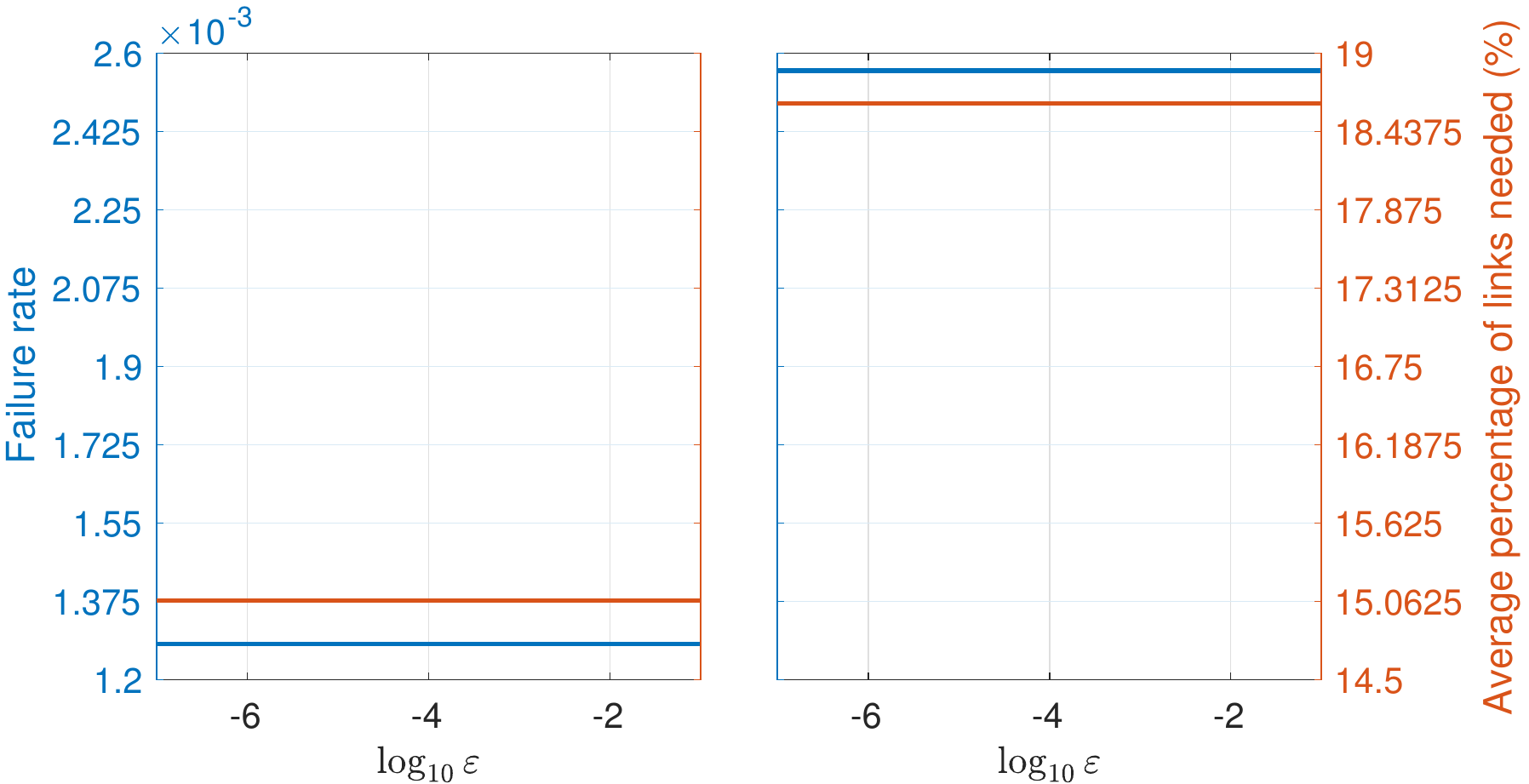}
		\caption{Failure rate and fraction of links needed for prediction vs $\epsilon$ for fixed $ \alpha=0.1$ if clustering by Origin/Destination (left) and Route (right).}
		\label{fig:leave1:eps}
\end{figure}
\begin{figure}[tbh]
	\centering
	\parbox[b]{.48\textwidth}{
		\centering
		\includegraphics[width=.48\columnwidth]{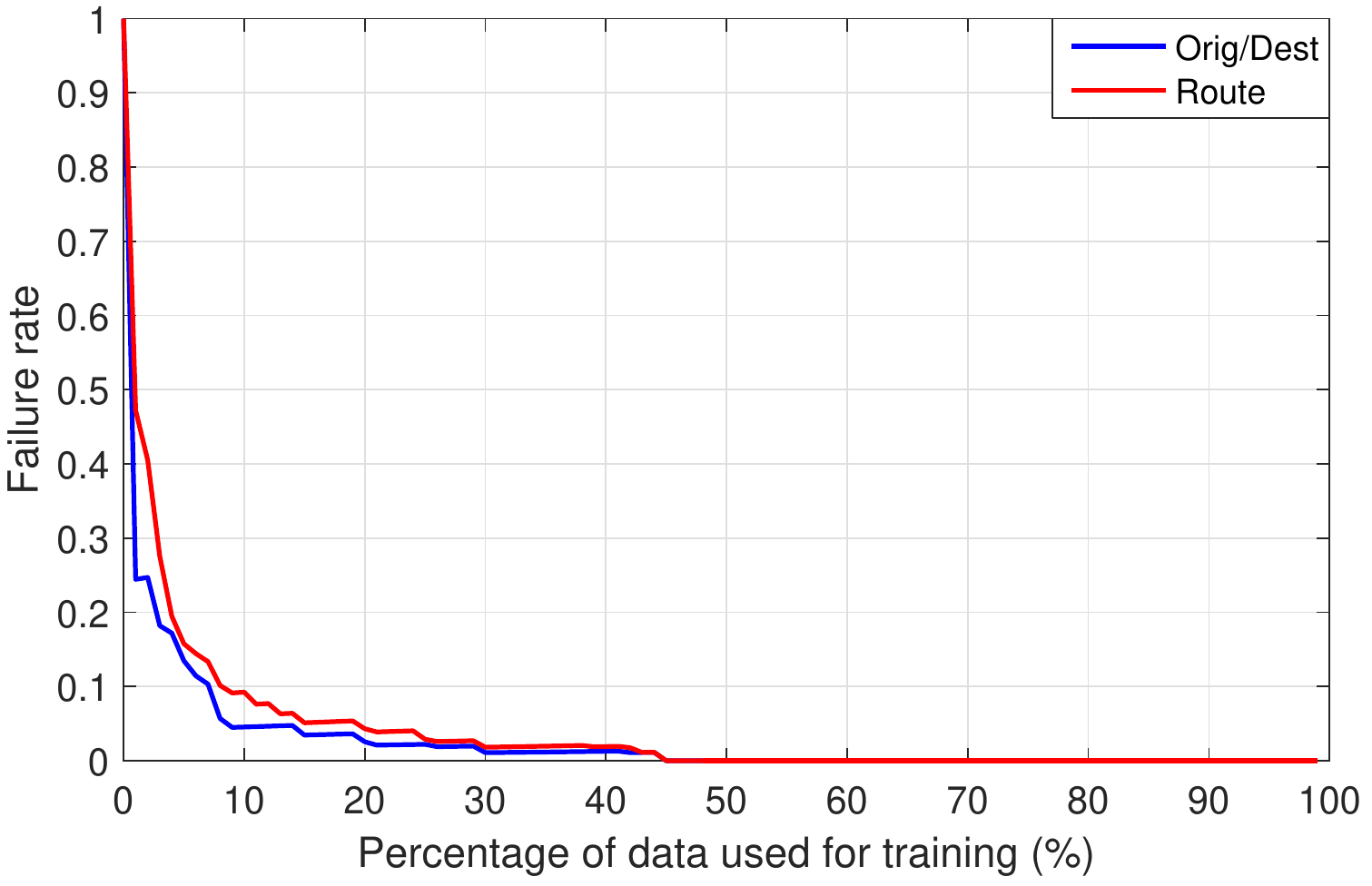}
		\caption{Decreasing failure rates as more trips are added to the training data for clustering by Origin/Destination and Route; $\alpha = 0.1$ and $\epsilon=10^{-6}$.}
		\label{fig:CaseCR1}
		}
	\hfill{}
	\parbox[b]{.48\textwidth}{
		\centering
		{\includegraphics[width=.48\columnwidth]{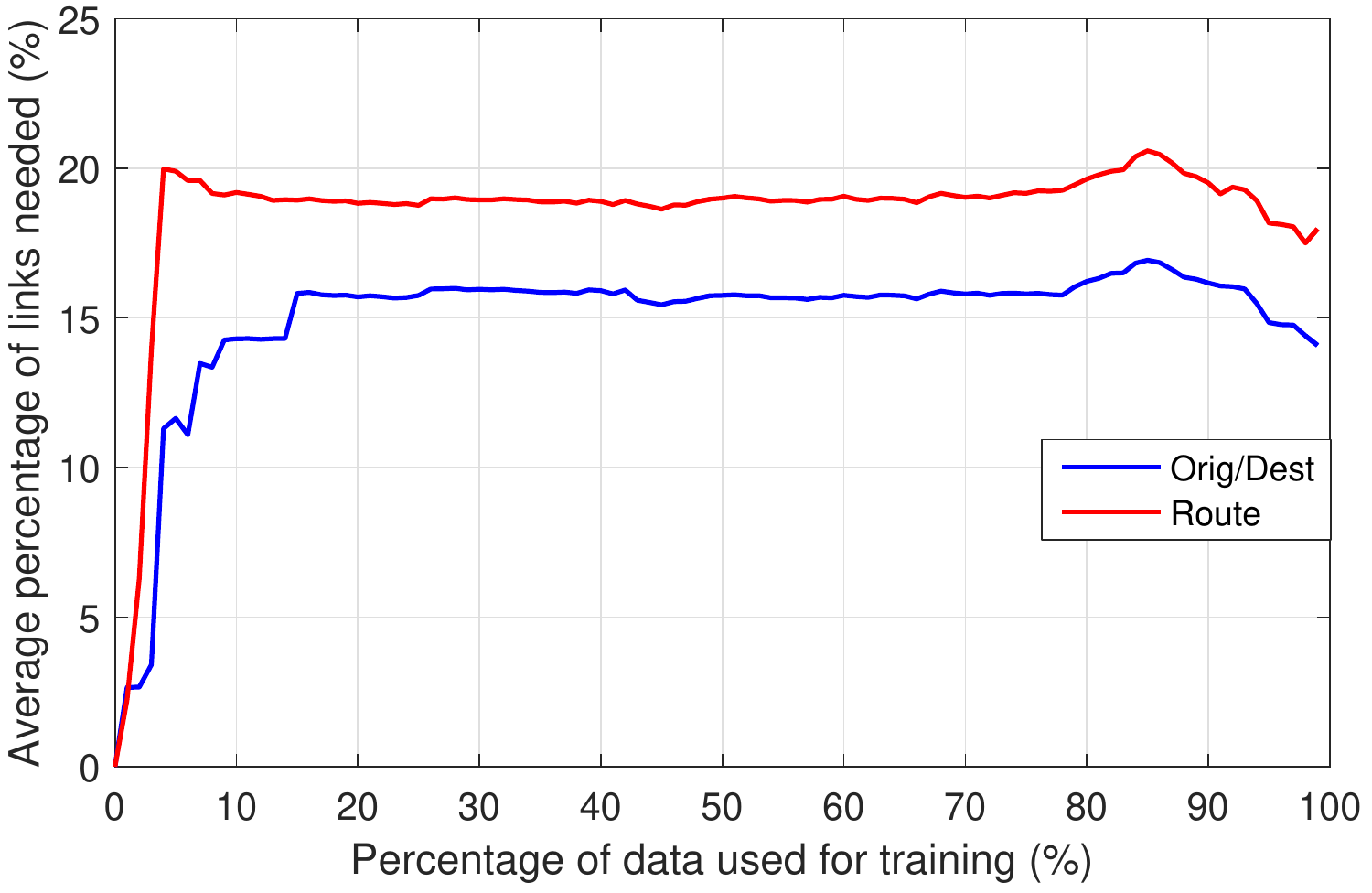}}
		\caption{Fraction of links needed as more trips are added to the training data for clustering by Origin/Destination and Route; $\alpha = 0.1$ and $\epsilon=10^{-6}$.}
		\label{fig:CaseCR2}
		}
\end{figure}
%
In all cases described below, we chose uniform probabilities for the prior $P(C_k)$ and initial probabilities $\pi^k$, i.e.\ $P(C_k)=1/N_C$ and $\pi^k_i=1/N$ $\forall k,i$. More careful choices could certainly improve prediction results, but as we shall see below, this simplest of choices is sufficient to demonstrate the efficacy of the algorithm.

We collected two quantities of interest: the \textit{failure rate} as the number of false predictions (which includes cases where the end of a trip is reached without a prediction being made) divided by the number of all trips predicted, and the \textit{fraction of trip completed} at the time of prediction, i.e.\ \#(road segments visited){/}\#(total road segments in trip). If a wrong prediction was made, we recorded the amount of links needed as \textsl{NaN}; these points are then excluded from the computation of averages.

In order to get an idea of good ranges for the parameters $\epsilon$ and $\alpha$, we performed a small initial \textbf{cross-validation} experiment by training on only 50\% of the data  and then predicting the route of the remaining 50\%. This was done for 8 random choices of training and testing set, on a grid of values $(\alpha,\epsilon)\in
\{10^{-4} ,   0.001  , 0.1,   0.2   , 0.25  ,  0.3  ,  0.35 ,   0.4\}\times\{10^{-7},10^{-5}, 0.001 ,   0.005,    0.01,    0.1\}$. 
 The results are shown in Fig.~\ref{fig:results} and indicate that the approach is robustly effective over a wide range of values for $\alpha$ and $\epsilon$:
with parameters in a reasonable range, e.g.\  $\alpha\in[0.01,0.1]$ and $\epsilon\in[10^{-7},10^{-5}]$, the routes are predicted accurately in more than 99\% of the test cases within the first 20\% of the trip. The results also show ``breakdown points:'' if the confidence level $\alpha$ is increased to the point where a posterior probability of 0.6 is sufficient for prediction already, the amount of false predictions increases rapidly; on the other hand, if the small probability parameter $\epsilon$ is chosen so large that it dominates the probabilities estimated from data, the amount of links necessary to distinguish routes increases rapidly. 

To investigate further, we then performed \textbf{leave-one-out cross-validation} along two ``slices'' on a finer grid: for each of the $N_H=781$ trips, we trained the Markov chains on the remaining $N_H-1=780$ trips; then, the trip that was left out was predicted. This was done for $\epsilon=10^{-6}$ and a finer grid of values for $\alpha$, and for $\alpha=0.1$ and a finer grid of values for $\epsilon$.  The entire procedure was repeated once for clustering by Origin/Destination, and once for clustering by Route. The results are shown in Figs.~\ref{fig:leave1:alpha:OD}-\ref{fig:leave1:eps}, and we observe:
\begin{itemize}
	\item The prediction accuracy is very robust with respect to $\alpha$, the failure rate is below 1\% for a wide range of $\alpha$, and declines rapidly once a breakdown point of $\alpha\approx0.25$ is reached.
	\item The same goes for the percentage of links needed to make a correct prediction. Additionally we note that, as should be expected, lower confidence in prediction (i.e.\ larger values of $\alpha$) tends to lead to fewer links needed for prediction; however this effect only kicks in once the failure rate increases quickly.
	\item Predicting the origin and destination appears to be slightly easier, since it consistently requires fewer links to do so. This is not surprising as there are only 17 clusters to choose from, whereas in the case of predicting the route there are 30.
	\item The robustness with respect to $\epsilon$ is even more pronounced: Fig.~\ref{fig:leave1:eps} indicates that, once a good value for $\alpha$ is selected, neither the failure rate nor the amount of links needed for prediction depend on $\epsilon$.
\end{itemize}

As a last experiment, we attempted to simulate the realistic situation of an in-car system improving its prediction model with each taken trip by \textbf{incrementally moving trips from the testing set to the training set}. Specifically, for the first data point, we trained on the first trip only and predicted the remaining $N_H-1=780$ ones. This obviously lead to an immediate wrong prediction of almost all trips. We then added the second trip, retrained and predicted the remaining 779 trips, and so on. The results are shown in Figs.~\ref{fig:CaseCR1} and~\ref{fig:CaseCR2}, and we see that once roughly 10\% of the trips have been taken (so around 75 trips), the algorithm predicts at least 90\% of the remaining trips correctly while needing on average between 15 and 20\% of the trip to be completed to make its prediction.
\section{CONCLUSIONS}
The contribution was twofold: on the one hand, the flexible framework of modelling route patterns as stochastic processes and using the associated likelihoods to update a posterior probability is introduced, and on the other, a concrete algorithm is presented, obtained by modelling the stochastic processes as Markov chains. 

The flexibility of the approach is only touched upon, there are many possible extensions which should be explored once a richer dataset is available {-- even though we worked on the generation of a realistic dataset (see Sec.\ \ref{ssec:data}), it is still a synthetic one and the excellent performance of the presented algorithm is hard to improve upon. Improvements to be explored using a more challenging, real dataset, include:} 
\begin{itemize}
		\item Other stochastic process models can be used. Indeed, the approach taken in~\cite{JulienRoutePred} fits into the outlined framework, if the choice of stochastic process is a naive Bayes model, i.e.\ if assuming that $P(r_1,\dotsc,r_L \given C_k) = \prod_{t=1}^L P(r_t\given C_k)$. We intend to test other stochastic process models, such as the recently developed closed-loop Markov modulated Markov chains~\cite{Epperlein2017} in the near future.
		\item So far, the available context is not used at all. For future practical applications however, the prior probabilities should be made dependent on such contextual variables as the day of the week or the time of the day, for instance by setting $P(C_k, \text{weekday}) \propto \frac{\text{\#(trips in $C_k$ that occurred on a weekday)}}{\text{\#(trips that occurred on a weekday)}}$, the prior probability if the current trip occurs on a weekday would be made proportional to the relative frequency of trips in $C_k$ among previous weekday trips; \cite{JulienRoutePred} has further details on context and its inclusion, only there, the contextual variables influence the stochastic process model directly instead of entering via the prior.
		\item The initial probabilities $\pi^k$ and small probabilities $\epsilon$ can be shaped to be larger for roads that are not on, but close to, the roads in cluster $C_k$, and smaller for roads that are far away. This can be expected to improve convergence of the posterior probabilities.
\end{itemize}
Overall, the success the approach has without tapping into such extensions is encouraging further research.

\bibliographystyle{siamplain}
\bibliography{destination-prediction}

\end{document}